\documentclass[letterpaper, 10 pt, conference]{ieeeconf}  %

\IEEEoverridecommandlockouts                              %

\usepackage[table]{xcolor}
\usepackage{hhline}
\usepackage{xspace}
\usepackage{algorithm}
\usepackage{booktabs}
\usepackage{graphicx}
\usepackage{algpseudocode}
\usepackage{siunitx}
\usepackage{svg}
\usepackage{float}
\usepackage{subcaption}
\usepackage{inconsolata}
\usepackage{xcolor}
\usepackage{bbding}
\usepackage{amsfonts}
\usepackage{wrapfig}
\usepackage{amsmath,bm}
\usepackage{multirow}
\usepackage{xcolor}
\usepackage{amsmath}
\usepackage{amssymb}
\usepackage{multirow}
\usepackage{booktabs}
\usepackage{graphicx}
\usepackage{mdframed}
\usepackage{caption}
\usepackage{verbatim}
\usepackage{xspace}
\usepackage{paralist}
\usepackage{cuted}

\usepackage[normalem]{ulem}
\usepackage{xcolor}

\usepackage[noadjust]{cite}

\usepackage{listings}
\usepackage{xcolor}  %

\usepackage{listings}
\definecolor{rblue}{rgb}{0,0.5,1}
\definecolor{awesome}{rgb}{1.0, 0.13, 0.32}
\definecolor{hollywoodcerise}{rgb}{0.96, 0.0, 0.63}
\definecolor{lasallegreen}{rgb}{0.03, 0.47, 0.19}
\definecolor{hanpurple}{rgb}{0.32, 0.09, 0.98}
\definecolor{green(pigment)}{rgb}{0.0, 0.65, 0.31}

\makeatletter
\let\NAT@parse\undefined
\makeatother
\usepackage[pagebackref=false,breaklinks=true,colorlinks,bookmarks=false]{hyperref}
\hypersetup{colorlinks=true,
    linkcolor={red},
    citecolor={hanpurple},
    urlcolor={magenta}
}

\title{\LARGE \bf
PanoAffordanceNet:\\Towards Holistic Affordance Grounding in 360{\textdegree} Indoor Environments
}
\author{Guoliang Zhu$^{1}$, Wanjun Jia$^{1}$, Caoyang Shao$^{1}$, Yuheng Zhang$^{1}$, Zhiyong Li$^{1,2}$, and Kailun Yang$^{1,2,\dag}$
\thanks{This work was supported in part by the National Natural Science Foundation of China (Grant No. 62473139), in part by the Hunan Provincial Research and Development Project (Grant No. 2025QK3019), and in part by the State Key Laboratory of Autonomous Intelligent Unmanned Systems (the opening project number ZZKF2025-2-10).}
\thanks{$^{1}$The authors are with the School of Artificial Intelligence and Robotics, Hunan University, Changsha 410012, China (email: kailun.yang@hnu.edu.cn).}%
\thanks{$^{2}$The authors are also with the National Engineering Research Center of Robot Visual Perception and Control Technology, Hunan University, Changsha 410082, China.}
\thanks{$^{\dag}$Corresponding author: Kailun Yang.}
}

\begin{document}

\maketitle
\thispagestyle{empty}
\pagestyle{empty}

\begin{abstract}
Global perception is essential for embodied agents in 360{\textdegree} spaces, yet current affordance grounding remains largely object-centric and restricted to perspective views. To bridge this gap, we introduce a novel task: Holistic Affordance Grounding in 360{\textdegree} Indoor Environments. This task faces unique challenges, including severe geometric distortions from Equirectangular Projection (ERP), semantic dispersion, and cross-scale alignment difficulties. We propose PanoAffordanceNet, an end-to-end framework featuring a Distortion-Aware Spectral Modulator (DASM) for latitude-dependent calibration and an Omni-Spherical Densification Head (OSDH) to restore topological continuity from sparse activations. By integrating multi-level constraints comprising pixel-wise, distributional, and region-text contrastive objectives, our framework effectively suppresses semantic drift under low supervision. Furthermore, we construct 360-AGD, the first high-quality panoramic affordance grounding dataset. Extensive experiments demonstrate that PanoAffordanceNet significantly outperforms existing methods, establishing a solid baseline for scene-level perception in embodied intelligence. The source code and benchmark dataset will be made publicly available at \url{https://github.com/GL-ZHU925/PanoAffordanceNet}.

\end{abstract}

\section{Introduction}

\begin{figure}[!t]
\centering
\includegraphics[width=0.48\textwidth]{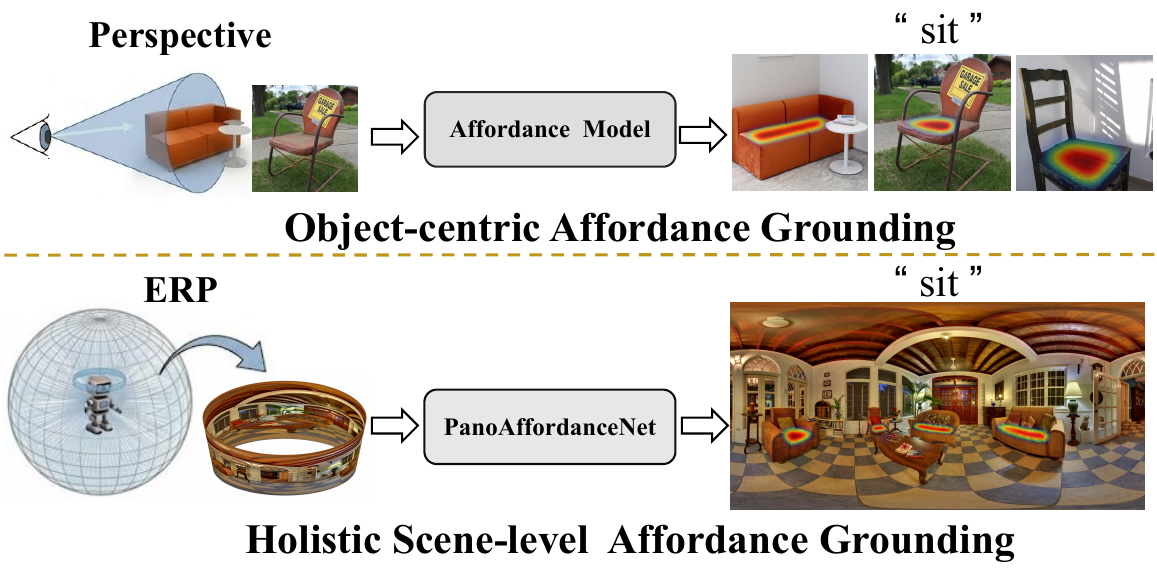}
\vskip-1ex
\caption{\small \textbf{Comparison of affordance grounding paradigms.} Traditional object-centric methods (\textbf{top}) are restricted to perspective views. Our proposed holistic scene-level affordance grounding (\textbf{bottom}) with PanoAffordanceNet enables omnidirectional functional perception in 360{\textdegree} indoor environments.  
}
\label{fig:teaser}
\vskip-4ex
\end{figure}

Endowing embodied agents (especially service robots) with the ability to perceive and interact with unstructured environments is one of the core pursuits in the field of artificial intelligence~\cite{duan2022survey}. 
Unlike traditional vision systems restricted to perspective views, robots inherently operate within a $360^{\circ}$ physical action
space~\cite{yu2025thinking,fried2018speaker,gattaux2025route}, which requires their perception systems to possess global awareness.
Among various perception tasks, identifying potential affordances in the environment~\cite{myers2015affordance} serves as a crucial bridge between visual perception and physical action planning~\cite{gibson2014theory}. 
However, existing visual affordance research is largely built upon an object-centric paradigm and focuses on functional understanding from restricted viewpoints, while largely overlooking the holistic spatial organization and geometric properties inherent in panoramic scenes. 
As illustrated in Fig.~\ref{fig:teaser}, this limited perspective creates a mismatch with the $360^{\circ}$ physical action space required for robots.
To this end, we advocate shifting the research focus toward a new task of holistic affordance grounding in $360^{\circ}$ indoor scenes (Fig.~\ref{fig:teaser}, bottom).

In recent years, visual affordance grounding has made significant progress, evolving from fully supervised methods to weakly supervised paradigms (\textit{e.g.}, LOCATE~\cite{li2023locate}, WSMA~\cite{xu2024weakly}), and further developing into foundation model and large language model-driven methods that utilize pseudo-labels, vision-language models, and open-vocabulary learning (\textit{e.g.}, OOAL~\cite{li2024one}, AffordanceLLM~\cite{qian2024affordancellm}). 
However, most of these advances rely on object-centric paradigms and are validated in simplified or controlled perspective-view scenarios.
When these models are directly extended to real panoramic scenes, their performance often drops precipitously due to high-entropy background interference, complex spatial compositions, and severe geometric distortions. 
This failure mode indicates that panoramic imagery 
fundamentally differs from perspective imagery; 
it inherently changes the distribution patterns of spatial features. 

In panoramic indoor scenes, achieving accurate affordance grounding remains challenged by three key factors. First, the Equirectangular Projection (ERP) introduces severe geometric distortion~\cite{coors2018spherenet}, especially near the polar regions, making it difficult for models to jointly preserve local interaction details and global functional structures. Second, non-uniform sampling~\cite{cao2024geometric} leads to highly sparse distributions of functional regions, resulting in scattered initial activations that are hard to aggregate into semantically coherent and boundary-consistent affordance areas. 
Third, due to the lack of dense pixel-level annotations, precisely aligning abstract affordance semantics with multi-scale regions in complex $360^{\circ}$ scenes is inherently difficult, often causing semantic drift.

To systematically address these challenges, we formulate the task of holistic affordance grounding in 360{\textdegree} indoor environments and propose  \textbf{PanoAffordanceNet}, an end-to-end one-shot learning framework specifically designed for $360^{\circ}$ indoor environments. 
Specifically, we design the Distortion-Aware Spectral Modulator (DASM) to isolate task-relevant geometric signals via dual-frequency spectral distillation, explicitly calibrating the severe geometric distortion and semantic dispersion caused by ERP at different latitudes. 
To address the signal fragmentation issue caused by non-uniform sampling in panoramas, we introduce a spherical-aware decoder leveraging its built-in Omni-Spherical Densification Head (OSDH). 
By utilizing spherical self-similarity and a seed-driven self-guidance mechanism, OSDH recovers sparse activations into topologically continuous complete functional areas. 
Finally, by integrating multi-level training objectives with pixel-level, distribution-level, and region-text contrastive constraints, the framework can effectively suppress semantic drift and ensure the robustness of global spatial localization under extremely low supervision. 
Meanwhile, we construct the first high-quality indoor panoramic affordance grounding dataset, \textbf{360-AGD}, to support standardized evaluation.
Extensive experiments show that our approach significantly outperforms state-of-the-art methods across all metrics.
Notably, it also maintains high competitiveness on the standard perspective AGD20K dataset~\cite{luo2022learning}, verifying its robustness across different domains.

To summarize, our contributions are as follows:
\begin{itemize}
    \item  We introduce a new task of holistic affordance grounding in $360^{\circ}$ indoor environments, shifting the paradigm from isolated object-level understanding to holistic scene-level reasoning.
    \item  We propose \textbf{PanoAffordanceNet}, which systematically addresses geometric distortion, sparse functional regions, and semantic drift through a Distortion-Aware Spectral Modulator (DASM), an Omni-Spherical Densification Head (OSDH), and a multi-level training objective, respectively.
    \item  We construct the first high-quality indoor panoramic affordance grounding dataset, \textbf{360-AGD}, and conduct extensive experiments to establish a benchmark and verify the effectiveness of our approach.
\end{itemize}

\section{Related Works}

\subsection{Visual Affordance Grounding}
Visual affordance grounding aims to localize interactive regions or action possibilities from visual observations, serving as a bridge between embodied perception and physical action planning. Early studies~\cite{do2018affordancenet} mainly relied on pixel-level supervision to learn object functional attributes, requiring large-scale annotated data.
To reduce annotation cost, subsequent works adopted weakly supervised paradigms using keypoints, image-level labels~\cite{sawatzky2017weakly}, or cross-view samples~\cite{luo2022learning}. These approaches further incorporated external interaction priors to transfer weak supervision into dense localization and alleviate supervision scarcity~\cite{luo2022learning,xu2024weakly}.

With the emergence of visual foundation models (VFMs), recent studies have explored data-efficient and open-vocabulary affordance grounding. Representative works exploit complementary priors (\textit{e.g.}, BiT-Align~\cite{huang2025resource}), selective contrastive learning~\cite{moon2025selective}, part-level semantics~\cite{xu2025weaklysupervised}, and segmentation transfer from SAM~\cite{jiang2025affordancesam}. As a typical one-shot approach, OOAL~\cite{li2024one} aligns part-level visual features with semantic embeddings for zero-shot generalization. However, most existing approaches are largely object-centric and evaluated in perspective-view settings, making them prone to semantic drift in panoramic scenes with complex layouts and multiple coexisting instances.

\subsection{Scene-Level Affordance Understanding}
As embodied tasks grow in complexity, affordance research is shifting from isolated object-level prediction toward holistic scene-level functional understanding. In the 2D vision domain, this transition is largely driven by foundation models, especially Large Vision-Language Models with rich world knowledge. WorldAfford~\cite{chen2024worldafford} explores scene-level grounding under natural language instructions using chain-of-thought reasoning to link abstract intentions with interaction regions, though it still depends on SAM-based object segmentation. Ramrakhya \textit{et al.}~\cite{ramrakhya2024seeingunseenvisualcommon} propose the semantic placement task, predicting plausible placement regions for unseen objects (\textit{e.g.}, pillows on a sofa) via visual common sense and semantic reasoning, enabling context-driven functional understanding beyond object-centric assumptions.

Parallel efforts in 3D vision explore scene-level affordance reasoning, including large-scale functional segmentation~\cite{delitzas2024scenefun3d}, intention-guided interaction grounding~\cite{liu2024grounding}, and multimodal embodied reasoning~\cite{wang2025affordbot}. Although 3D methods provide precise geometric constraints, they suffer from high annotation costs and the lack of mature 3D foundation models. In contrast, panoramic imagery offers a promising intermediate representation by combining the semantic generalization of 2D foundation models with $360^{\circ}$ omnidirectional spatial context for embodied decision-making.

\subsection{Indoor Panoramic Perception}
Perceiving $360^{\circ}$ panoramic images faces inherent challenges such as geometric distortion, non-uniform sampling, and cross-boundary semantic discontinuity caused by Equirectangular Projection (ERP)~\cite{coors2018spherenet}. Prior work on semantic segmentation~\cite{zhang2022bending,cao2024geometric} and object detection~\cite{coors2018spherenet} mitigates these issues through geometry-aware or spherical convolutions~\cite{cao2024geometric,coors2018spherenet} and global context priors~\cite{dong2024panocontext}. However, these approaches target generic recognition and overlook the compounded effect of panoramic distortion on fine-grained affordance localization. In affordance grounding, sparse supervision and multi-scale propagation are particularly vulnerable to geometric deformation. This motivates affordance grounding tailored to $360^{\circ}$ indoor environments.

\begin{figure*}[t!]
\centering
\includegraphics[width=0.98\textwidth]{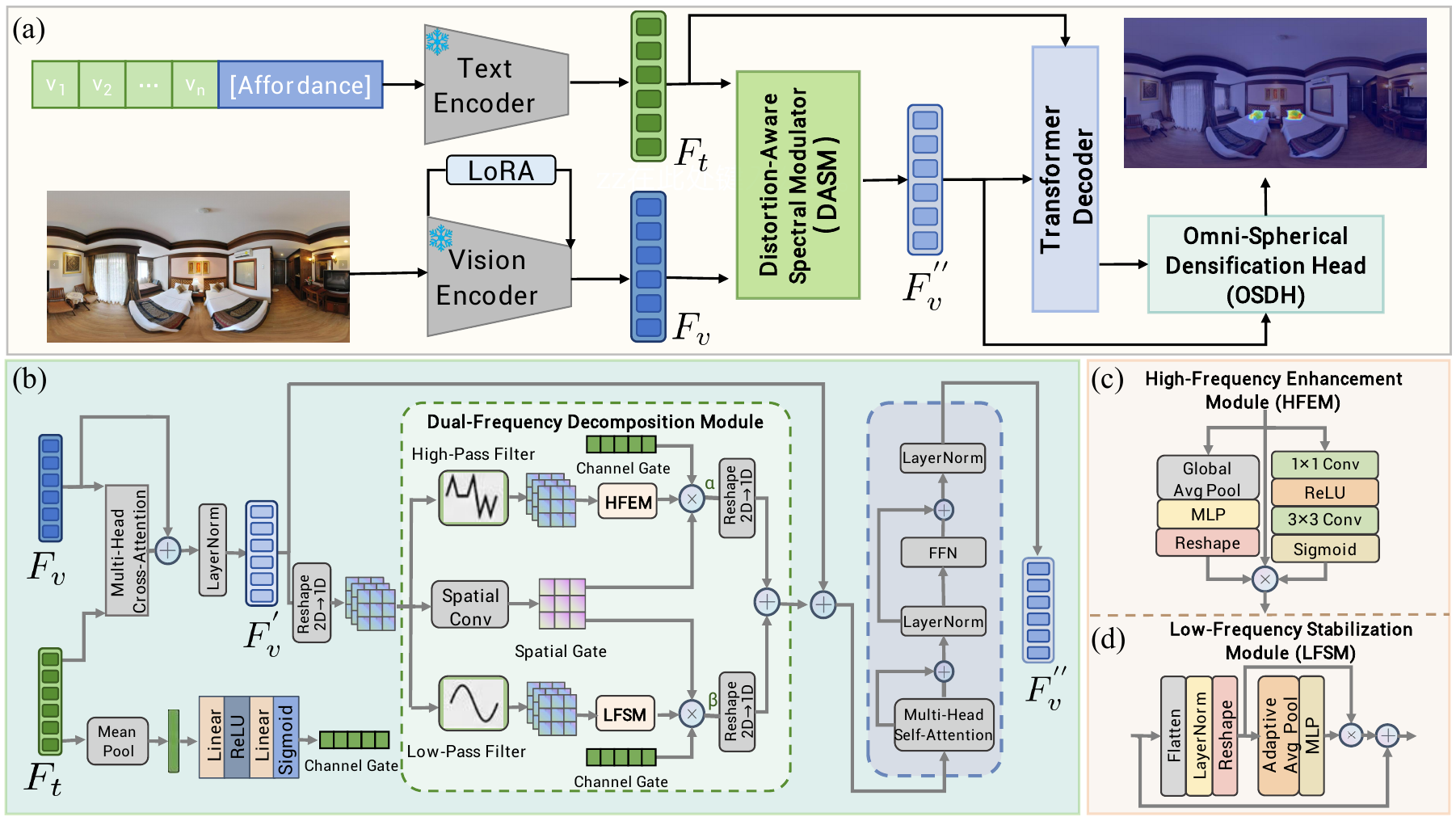} 
\vskip-1ex
\caption{\small \textbf{Overview of PanoAffordanceNet.}
(a) Parameter-efficient dual-encoder framework with distortion-aware modulation and spherical densification. 
(b) Distortion-Aware Spectral Modulator (DASM) for latitude-adaptive frequency decomposition.
(c) HFEM and (d) LFSM for interaction boundary sharpening and structural stabilization, respectively.
}
\label{fig:framework}
\vskip-3ex
\end{figure*}

\section{Method}

\subsection{Overview} 
In this section, we present \textbf{PanoAffordanceNet}, an end-to-end framework for one-shot affordance grounding in $360^{\circ}$ indoor environments. Due to geometric distortion, semantic dispersion, and scarce supervision, panoramic affordance grounding is highly challenging. Accordingly, we propose a modular pipeline, as illustrated in Fig.~\ref{fig:framework}, consisting of:
(1) dual-encoder feature extraction with LoRA-based parameter-efficient adaptation for multi-modal representation learning (Sec.~\ref{sec:backbone});
(2) the Distortion-Aware Spectral Modulator (DASM), which utilizes dual-frequency spectral distillation to isolate task-pertinent geometric signals and mitigate signal distribution distortions caused by ERP at varied latitudes (Sec.~\ref{sec:dasm}); 
(3) a Spherical-Aware Hierarchical Decoder (Sec.~\ref{sec:decoder}) that facilitates global semantic discovery for initial anchoring, followed by the Omni-Spherical Densification Head (OSDH) to recover complete functional regions that are geometrically coherent and topologically continuous; and 
(4) a Multi-level Training Objective incorporating pixel-level accuracy, distributional topology consistency, and region-text contrastive alignment (Sec.~\ref{sec:loss}).

\subsection{Feature Extraction} 
\label{sec:backbone}

We construct our multi-modal foundation from two pre-trained encoders, efficiently adapted to the one-shot panoramic grounding task. For visual perception, we employ DINOv2 (ViT-B/14)~\cite{oquab2023dinov2} to extract patch-level features $\mathbf{F}_v \in \mathbb{R}^{B \times L \times D}$.
To mitigate overfitting under sparse one-shot annotations while accommodating ERP-induced geometric distortions, we apply Low-Rank Adaptation (LoRA)~\cite{hu2021lora} by inserting trainable low-rank matrices into the attention layers of the Transformer. 
For linguistic guidance, the pre-trained CLIP text encoder (ViT-B/16)~\cite{radford2021learning} is combined with a CoOp-based prompt learner~\cite{zhou2022learning}, producing context-aware text embeddings $\mathbf{F}_t \in \mathbb{R}^{B \times C \times D}$ for the $C$ affordance classes. 
These multi-modal features provide a robust semantic foundation for subsequent modulation and grounding stages.

\subsection{Distortion-Aware Spectral Modulator }\label{sec:dasm}

Equirectangular Projection (ERP) images offer a full $360^{\circ}$ view~\cite{yu2025thinking} but suffer from latitude-dependent geometric distortion (especially near the poles) and semantic dispersion from background clutter, weakening functional cues. To address this, we introduce a Distortion-Aware Spectral Modulator (DASM), as illustrated in Fig.~\ref{fig:framework}(b). DASM isolates task-relevant geometry via dual-frequency spectral distillation while compensating for ERP-induced artifacts.

The modulation process begins with cross-modal semantic injection. 
Given visual features $\mathbf{F}_v \in \mathbb{R}^{L \times D}$ and text embeddings $\mathbf{F}_t \in \mathbb{R}^{C \times D}$ from the backbone, we inject linguistic guidance via a multi-head attention mechanism:
\begin{equation}
\mathbf{F}'_v = \text{Softmax}\big( (\mathbf{F}_v \mathbf{W}_Q)(\mathbf{F}_t \mathbf{W}_K)^\top / \sqrt{d} \big) (\mathbf{F}_t \mathbf{W}_V),
\end{equation}
where {\small $\mathbf{W}_{\{Q,K,V\}}$} are learnable projections. 
This stage activates semantically relevant regions on the visual manifold.

Subsequently, $\mathbf{F}'v$ is reshaped into a spatial feature map and decoupled into high- and low-frequency components :
\begin{equation}
\mathbf{F}_h = \nabla^2 * \mathbf{F}'_v, \quad \mathbf{F}_l = \mathcal{K}_{\sigma} * \mathbf{F}'_v,
\end{equation}
where the low-frequency component is obtained via Gaussian smoothing with kernel $\mathcal{K}{\sigma}$, while the high-frequency component is extracted using a Laplacian operator $\nabla^2$ to emphasize boundary and interaction contours.

To address ERP-induced frequency imbalance, where sharp edges are preserved near the equator but structures are stretched at the poles, we apply targeted compensation to each branch:
the High-Frequency Enhancement Module (HFEM, Fig.~\ref{fig:framework}(c)) sharpens the interaction boundaries in equatorial regions while suppressing artifacts amplified at the poles;
the Low-Frequency Stabilization Module (LFSM, Fig.~\ref{fig:framework}(d)) maintains global structural consistency near the poles to mitigate semantic fragmentation from stretching.

The refined branches are selectively fused via hybrid gated modulation mechanism, in which a language-driven channel gate $\mathbf{g}_{\text{ch}}$ emphasizes task-relevant semantics across channels, while a self-adaptive spatial gate $\mathbf{g}_{\text{sp}}$, derived from the feature map itself, anchors salient regions within the 360$^\circ$ field.
The fused features are obtained through gated residual addition:
\begin{equation}
\mathbf{F}_{\text{freq}} = \mathbf{F}'_v + \sum_{k \in \{h,l\}} \lambda_k \, (\mathbf{g}_{\text{ch}} \odot \mathbf{g}_{\text{sp}} \odot \mathbf{F}_k),
\end{equation}
with learnable scalars $\lambda_k$.

Finally, a contextual re-aggregation stage (Fig.~\ref{fig:framework}b, blue box) employs MHSA and FFN to exchange global context across the panoramic field, restoring spatial coherence and yielding a distortion-robust, affordance-aware feature $\mathbf{F}''_v$. 

\subsection{Spherical-Aware Hierarchical Decoder}\label{sec:decoder}
In $360^{\circ}$ panoramic scenes, affordance cues are sparse and unevenly distributed due to ERP’s non-uniform sampling, leading to fragmented and spatially disconnected predictions, especially near the poles.
To address this issue, we propose a spherical-aware hierarchical decoder that performs global semantic anchoring and densifies sparse signals into topologically continuous regions via the Omni-Spherical Densification Head (OSDH).

\textbf{Global Semantic Discovery.} 
A lightweight transformer decoder uses task-adaptive text embeddings $\mathbf{F}_t$ as queries to cross-attend to the refined visual features $\mathbf{F}''_v$. The updated queries generate initial affordance maps $\mathbf{A}_{\text{init}} \in \mathbb{R}^{B \times C \times L}$ via scaled dot-product similarity:
\begin{equation}
\mathbf{A}_{\text{init}} = (\mathbf{F}_t \cdot {\mathbf{F}''_v}^\top) \cdot D^{-0.5},
\end{equation}
where $D^{-0.5}$ stabilizes attention variance. 
Although $\mathbf{A}_{\text{init}}$ captures potential affordance regions effectively, it remains inherently fragmented and lacks the geometric continuity required for precise interaction grounding.

\begin{figure}[!t]
\centering
\includegraphics[width=0.48\textwidth]{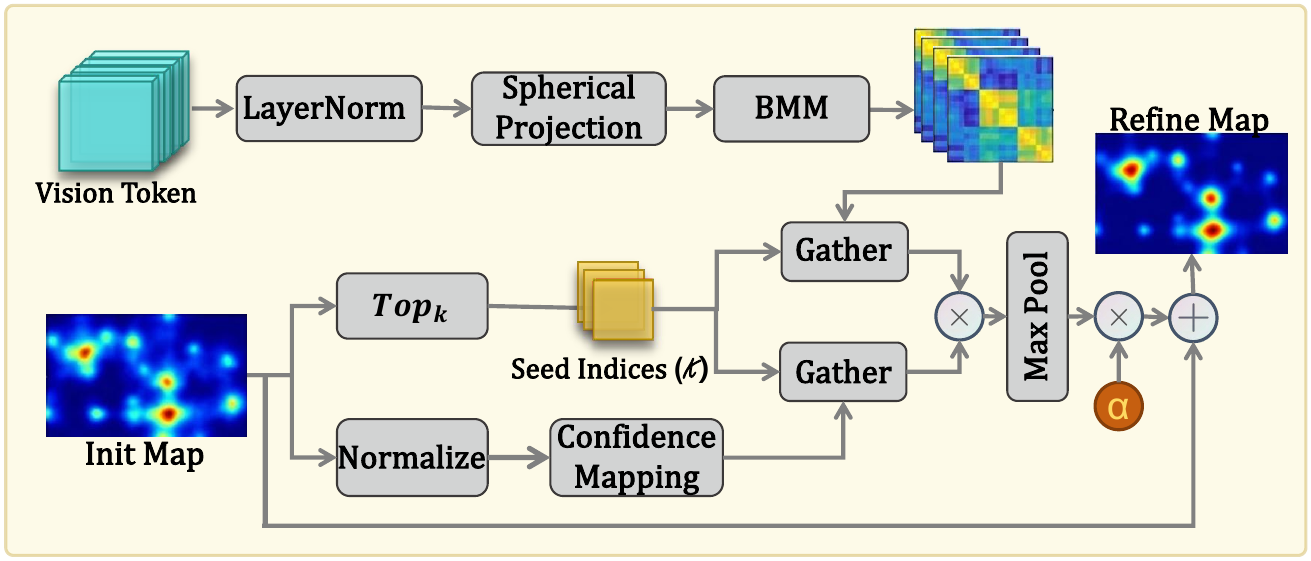}
\vskip-1ex
\caption{\small \textbf{Architecture of the Omni-Spherical Densification Head (OSDH).} Visual tokens undergo spherical projection to construct a cosine affinity matrix. Sparse initial activations are then densified via top-$k$ seed selection, confidence-guided noise suppression, and max propagation with a learnable residual scalar $\alpha$.}
\label{fig:OSDH}
\vskip-3ex
\end{figure}

\textbf{Omni-Spherical Densification Head.} 
To restore topological continuity on the spherical manifold, OSDH leverages visual self-similarity as a structural inductive bias (detailed in Fig.~\ref{fig:OSDH}).
Specifically, we project the refined visual features $\mathbf{F}''_v$ onto the unit hypersphere and construct a symmetric affinity matrix $\mathcal{S} \in \mathbb{R}^{L \times L}$ via cosine similarity:
\begin{equation}
\mathcal{S}_{ij} = (\mathbf{f}''_{v,i} \cdot \mathbf{f}''_{v,j})/(\|\mathbf{f}''_{v,i}\| \|\mathbf{f}''_{v,j}\|).
\end{equation}
Simultaneously, high-confidence seeds $\mathcal{K}$ are selected via top-$k$. Spurious noise is suppressed with a confidence map:
\begin{equation}
\mathcal{C} = \mathrm{Sigmoid}\!\left( (\mathbf{A}_{\text{init}} - \mu_{\mathbf{A}})/(\sigma_{\mathbf{A}}/T) \right),
\end{equation}
The final densified map is obtained via seed propagation:
\begin{equation}
\mathbf{A}_{\text{refined}} = \mathbf{A}_{\text{init}} + \alpha \cdot \max_{j \in \mathcal{K}} (\mathcal{S}_{ij} \cdot \mathcal{C}_j),
\end{equation}
where $\alpha$ is a learnable residual scalar. 
OSDH efficiently recovers complete, coherent regions for panoramic geometry.

This design seamlessly integrates global semantic alignment with spherical structural propagation, producing dense and continuous affordance predictions in panoramic scenes.

\begin{figure*}[t!]
\centering
\includegraphics[width=0.99\textwidth]{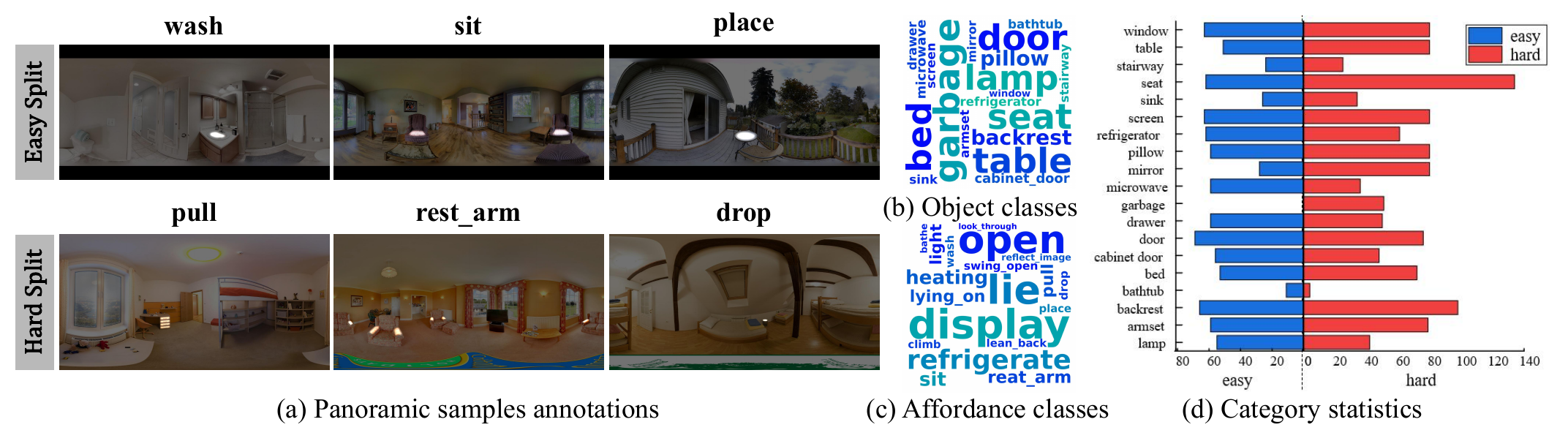} 
\vskip-1ex
\caption{\small \textbf{Properties of the 360-AGD dataset.} 
(a) Representative examples from the dataset. 
(b) Word cloud of object categories.
(c) Word cloud of affordance categories. 
(d) Statistical distribution of image counts per affordance across the Easy and Hard splits.}
\label{fig:dataset}
\vskip-3ex
\end{figure*}

\subsection{Multi-Level Training Objective}\label{sec:loss}
Indoor panoramic environments exhibit large-scale variation and multiple affordances per object (\textit{e.g.}, distinguishing ``grasp'' on a sofa armrest from ``sit'' on the cushion), requiring robust multi-scale reasoning and precise region–semantic alignment. After mitigating distortion and sparsity, we impose multi-level training objectives at the pixel, distribution, and region–text levels to suppress semantic drift and ensure robust global grounding.

Pixel-wise localization is supervised using a Binary Cross-Entropy loss ($\mathcal{L}_{BCE}$) to ensure accurate activation. To preserve panoramic topological continuity, a KL divergence loss ($\mathcal{L}_{KL}$) is introduced to enforce the predicted heatmap $\mathcal{M}$ to approximate the ground-truth distribution $\hat{\mathcal{M}}$:
\begin{equation}
\mathcal{L}_{KL} = \sum_{i,j} \hat{\mathcal{M}}_{i,j} \log \big( (\hat{\mathcal{M}}_{i,j} + \epsilon) / (\mathcal{M}_{i,j} + \epsilon) \big),
\end{equation}
thus maintaining global consistency in shape and intensity.

However, $\mathcal{L}_{BCE}$  and $\mathcal{L}_{KL}$ alone cannot resolve semantic ambiguity, particularly when multiple affordance attributes exist within the same object. Therefore, we introduce an InfoNCE-based Region--Text Contrastive loss ($\mathcal{L}_{RTC}$) to establish semantic correspondence between visual regions and affordance concepts. The densified visual features $\mathbf{F}''_v$ are pooled into a region-level representation:
\begin{equation}
\mathbf{v}_c = \textstyle \left( \sum_{l=1}^L \hat{M}_{c,l} \mathbf{f}''_{v,l} \right) \big/ \left( \sum_k \hat{M}_{c,k} + \epsilon \right),
\end{equation}
which is aligned with the corresponding text embedding:
\begin{equation}
\mathcal{L}_{RTC} = -\frac{1}{C} \sum_{c=1}^{C} \log \frac{\exp(\text{sim}(\mathbf{v}_c,\mathbf{t}_c)/\tau)}{\sum_{k=1}^{C} \exp(\text{sim}(\mathbf{v}_c,\mathbf{t}_k)/\tau)}.
\end{equation}
This semantic constraint significantly alleviates misalignment across regions and affordance attributes, allowing language supervision to accurately ground to specific visual areas.

The final training objective is formulated as:
\begin{equation}
\mathcal{L}_{total} = \lambda_1 \mathcal{L}_{BCE} + \lambda_2 \mathcal{L}_{KL} + \lambda_3 \mathcal{L}_{RTC},
\end{equation}
where $\lambda_1, \lambda_2$, and $\lambda_3$ are hyperparameters used to balance the contribution of each loss component.

\section{360-AGD: Established Dataset}

Existing affordance datasets~\cite{luo2022learning,liu2024grounding} mainly target standard-view 2D images or 3D point clouds, and no public benchmark currently supports interactive affordance grounding in panoramic indoor scenes. 
To bridge this gap, we introduce \textbf{360-AGD}, the first dataset with precise annotations of interaction regions in panoramic images.
This benchmark is designed to evaluate model generalization and robustness across indoor environments of diverse complexity.

\textbf{Data Collection.} 
To build a benchmark capable of comprehensively evaluating model performance, we collect images from several existing large-scale panorama datasets and organize them into an ``Easy Split'' and a ``Hard Split''. 
The Easy Split is designed to evaluate the model's fundamental performance in relatively clean and simple indoor scenes. 
Its images are primarily sourced from the 360-Indoor~\cite{chou2020360} and Gibson~\cite{xia2018gibson} datasets, with original resolutions of approximately $512{\times}1024$. 
We further establish a Hard Split to test generalization across more challenging environments. Sourced from PanoContext~\cite{zhang2014panocontext} and Sun360~\cite{xiao2012recognizing}, these samples exhibit superior visual complexity and high-fidelity details, with original resolutions reaching $4552{\times}9104$.
In total, the Easy and Hard splits contain approximately $800$ and $1,200$ panoramic images, respectively.

\textbf{Data Annotation.}
Our annotation task targets $19$ affordance classes within complex, $360^{\circ}$ panoramic indoor scenes. 
A key challenge is that a single affordance (\textit{e.g.}, sit, lie) can correspond to multiple, spatially disjoint regions within the same scene. 
Consequently, annotators are instructed to identify and mark all visible instances of each affordance class. However, given the visual clutter and complexity of these scenes, we ignore heavily occluded interaction regions whose boundaries are ambiguous. To efficiently capture such multi-region affordances without costly pixel-wise segmentation, we adopt a keypoint-based supervision strategy. Annotators place multiple keypoints within the boundaries of all valid (non-occluded) interaction areas.
These keypoint sets are then transformed into a continuous ground-truth signal by applying a Gaussian kernel to each point, generating a probability heatmap for each affordance class. Some annotation examples are shown in Fig.~\ref{fig:dataset}(a).

%---------------------------
\begin{table*}[t]
\centering
\caption{\small Quantitative comparison across different domains. 
\textbf{Top}: Performance on our proposed 360-AGD dataset. 
\textbf{Bottom}: Generalization results on the perspective AGD20K dataset~\cite{luo2022learning}.}
\label{tab:main_results}

\footnotesize
\renewcommand{\arraystretch}{1.18} 
\setlength{\tabcolsep}{6pt}        

\begin{tabular*}{\linewidth}{
@{\extracolsep{\fill}}
                        >{\raggedright\arraybackslash}p{2.5cm} 
                          >{\centering\arraybackslash}p{3.0cm}       
                          | >{\centering\arraybackslash}p{1.5cm}     
                          >{\centering\arraybackslash}p{1.5cm}     
                          >{\centering\arraybackslash}p{1.5cm} 
                          | >{\centering\arraybackslash}p{1.5cm}
                          >{\centering\arraybackslash}p{1.5cm}
                          >{\centering\arraybackslash}p{1.5cm}}
\toprule
\rowcolor[gray]{.95} \multicolumn{8}{c}{\textbf{(a) Evaluation on the Proposed 360-AGD Dataset }} \\ \midrule
\multirow{2}{*}{\textbf{Method}} & \multirow{2}{*}{\textbf{Supervision}} 
& \multicolumn{3}{c|}{\textbf{Easy Split}} & \multicolumn{3}{c}{\textbf{Hard Split}} \\[-0.04em]
\cmidrule(lr){3-5} \cmidrule(lr){6-8}
& & \textbf{KLD$\downarrow$} & \textbf{SIM$\uparrow$} & \textbf{NSS$\uparrow$} & \textbf{KLD$\downarrow$} & \textbf{SIM$\uparrow$} & \textbf{NSS$\uparrow$} \\
\midrule
OOAL~\cite{li2024one} & One-shot & 2.868 & 0.117 & 1.267 & 3.067 & 0.097 & 1.484 \\
OS-AGDO~\cite{jia2025one} & One-shot &2.853 & 0.124 &1.299 & 2.965 & 0.115 & 1.484 \\
\textbf{Ours} & One-shot &  \textbf{1.270} & \textbf{0.506} & \textbf{4.490}& \textbf{1.306} & \textbf{0.474} & \textbf{4.398} \\
\bottomrule
\end{tabular*}

\vspace{0.2cm} 

\begin{tabular*}{\linewidth}{
@{\extracolsep{\fill}}
>{\raggedright\arraybackslash}p{2.5cm} 
                          >{\centering\arraybackslash}p{3.0cm}
                          | >{\centering\arraybackslash}p{1.5cm}
                          >{\centering\arraybackslash}p{1.5cm}
                          >{\centering\arraybackslash}p{1.5cm}
                          | >{\centering\arraybackslash}p{1.5cm}
                          >{\centering\arraybackslash}p{1.5cm}
                          >{\centering\arraybackslash}p{1.5cm}}
\toprule
\rowcolor[gray]{.95} \multicolumn{8}{c}{\textbf{(b) Generalization Results on the Perspective AGD20K Dataset}} \\ \midrule
\multirow{2}{*}{\textbf{Method}} & \multirow{2}{*}{\textbf{Supervision}} 
& \multicolumn{3}{c|}{\textbf{Seen Split}} & \multicolumn{3}{c}{\textbf{Unseen Split}} \\[-0.04em]
\cmidrule(lr){3-5} \cmidrule(lr){6-8}
& & \textbf{KLD$\downarrow$} & \textbf{SIM$\uparrow$} & \textbf{NSS$\uparrow$} & \textbf{KLD$\downarrow$} & \textbf{SIM$\uparrow$} & \textbf{NSS$\uparrow$} \\
\midrule
LOCATE~\cite{li2023locate} & Weakly & 1.226 & 0.401 & 1.177 & 1.405 & 0.372 &1.157 \\
WSMA~\cite{xu2025weaklysupervised} & Weakly & 1.176 &0.416 &1.247 &1.335& 0.382& 1.220\\
 LoopTrans~\cite{tang2025closed} &Weakly & 1.088 & 0.445 &1.322& 1.247 &0.403 &1.315\\
OS-AGDO~\cite{jia2025one} & One-shot & 1.320 & 0.390 & 1.021& 1.398 & 0.382 &1.174 \\
OOAL~\cite{li2024one} & One-shot & 0.740 & 0.577 & 1.745 & 1.070 &0.461 &1.503\\
\textbf{Ours} & One-shot & \textbf{0.739} & \textbf{0.616} & \textbf{1.750} & 1.185 & \textbf{0.475} & 1.419 \\
\bottomrule
\end{tabular*}
\vskip-3ex
\end{table*}

\textbf{Statistic Analysis.}
To provide a comprehensive understanding of the 360-AGD dataset, we analyze its key characteristics across multiple dimensions. Fig.~\ref{fig:dataset}(b) and  Fig.~\ref{fig:dataset}(c) present the word clouds for object and affordance categories, respectively, highlighting the semantic diversity of the indoor scenes. 
Furthermore,  Fig.~\ref{fig:dataset}(d) provides a detailed statistical breakdown of affordance distributions across the Easy and Hard splits.
The distinct data distributions across these levels further reveal the inherent complexity of panoramic environments, posing a challenge for robust affordance grounding.

\section{Experiments}

\subsection{Experimental Setup}

\textbf{Implementation Details} 
The visual encoder adopts a DINOv2-Base model~\cite{oquab2023dinov2} pre-trained on ImageNet, while the text encoder uses a pre-trained CLIP model~\cite{radford2021learning}. 
The model is trained end-to-end using the AdamW optimizer with an initial learning rate of $1e{-}5$ and a cosine annealing schedule.
Training is performed on two NVIDIA A6000 GPUs for a total of $20k$ iterations with a batch size of $4$. 
All panoramic inputs are uniformly resized to a resolution of $560{\times}1120$. 
To enhance robustness, in addition to standard random flipping and color jittering, we introduce panoramic-specific data augmentations, including random rotations of $\pm 3^\circ$, random scaling by $\pm 5\%$, and horizontal wraparound shifts,  tailored to enforce the rotational invariance inherent to the $360^\circ$ topology.
Furthermore, during training, a Gaussian blur is then applied to each binary annotation mask to obtain the final soft supervision heatmaps.

\textbf{Evaluation Metrics.} 
Following established practices for affordance grounding evaluation, we assess our model's performance using three standard metrics: Kullback-Leibler Divergence (KLD), Similarity (SIM), and Normalized Scanpath Saliency (NSS).
KLD quantifies the distributional difference between the predicted and ground-truth heatmaps (lower is better). 
SIM measures their histogram intersection, and NSS calculates the normalized prediction score at ground-truth keypoint locations.
For both SIM and NSS, higher values indicate better performance.

\subsection{Comparison Results}
As the first framework dedicated to one-shot affordance grounding in panoramic environments, PanoAffordanceNet is benchmarked against two leading 2D one-shot methods, OOAL~\cite{li2024one} and OS-AGDO~\cite{jia2025one}, both of which have been adapted for equirectangular projections under the same one-shot protocol. 
As summarized in Table~\ref{tab:main_results}(a), our framework consistently demonstrates superiority across all metrics for both Easy and Hard splits, establishing a robust performance baseline for this specialized task.
Furthermore, generalization experiments on the perspective AGD20K dataset~\cite{luo2022learning} (Table~\ref{tab:main_results}(b)) show that PanoAffordanceNet achieves strong performance on the Seen split and remains competitive on Unseen categories. This stems from the panoramic-specific modules, which produce smoother affordance distributions. While this slightly degrades point-sensitive metrics, it improves SIM and validates robust cross-view generalization.

\begin{table}[!t]
    \centering
    \setlength{\tabcolsep}{7pt}  
    \caption{\small Ablation study of model components on the 360-AGD Hard Split.}
    \label{tab:module_ablation}
    \vskip-1ex
    \begin{tabular}{ccc|ccc}
        \toprule
        \textbf{LoRA} & \textbf{DASM} & \textbf{OSDH} & \textbf{KLD $\downarrow$} & \textbf{SIM $\uparrow$} & \textbf{NSS $\uparrow$} \\
        \midrule
             &           &            &  1.475  &  0.416  &  4.196  \\ %
        \checkmark &     &            &  1.421  &  0.429  & 4.257  \\ %
        \checkmark &     &  \checkmark  & 1.380  &  0.450  & 4.317 \\ %
        \checkmark &  \checkmark  &  &  1.359  &  0.448 & 4.339 \\ %
        \checkmark & \checkmark & \checkmark 
        &  \textbf{1.306} &  \textbf{0.474}  &  \textbf{4.398}  \\ %
        \bottomrule
    \end{tabular}
    \vskip-5ex
\end{table}

\subsection{Performance Analysis}
\begin{figure*}[t!]
  \centering
  \vskip-1ex
  \includegraphics[width=0.98\textwidth]{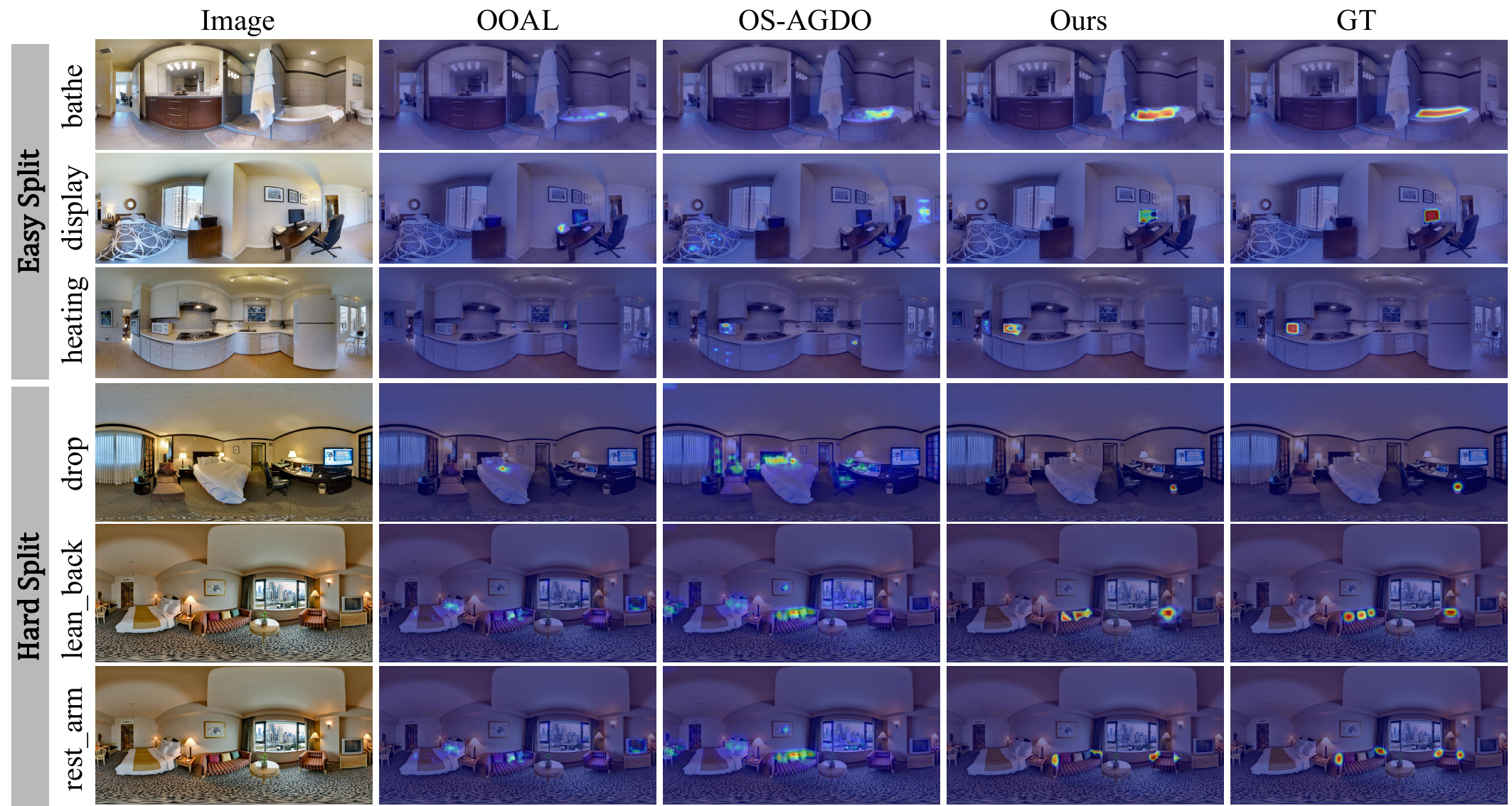}
  \caption{\small \textbf{Qualitative comparison} between the proposed PanoAffordanceNet and state-of-the-art one-shot affordance grounding methods, including OOAL~\cite{li2024one} and OS-AGDO~\cite{jia2025one}, on the established 360-AGD dataset.}
  \vskip-3ex
  \label{fig:visualization}
\end{figure*}

To further elucidate the underlying mechanisms of these performance discrepancies, we conduct a comparative visual analysis of the predicted affordance heatmaps across diverse cluttered indoor scenes (Fig.~\ref{fig:visualization}).
Conventional 2D perspective-based paradigms undergo a catastrophic breakdown of spatial reasoning when directly applied to 360-degree equirectangular inputs.
The heatmaps generated by OOAL~\cite{li2024one} and OS-AGDO~\cite{jia2025one} are characterized by erratic, fragmented activations and a pervasive lack of structural coherence, frequently exhibiting severe semantic drift where high-response regions systematically deviate from the ground-truth functional areas.
In contrast, PanoAffordanceNet yields cleaner and more precisely localized heatmaps with superior spatial integrity. 
As shown in the last two rows of Fig.~\ref{fig:visualization}, for different tasks on the same object within the same scene such as``\textit{lean\_back}'' and ``\textit{rest\_arm}'', the response maps generated by the other two methods exhibit high similarity and semantic confusion. In contrast, our method achieves more precise cross-modal alignment, enabling the generation of clearly differentiated predictions.

\subsection{Ablation Studies}
To assess the contribution of each component in PanoAffordanceNet, we perform ablation on the 360-AGD Hard Split under a unified one-shot setting for fair comparison.

\begin{table}[!t]
  \centering
  \setlength{\tabcolsep}{10pt}   
  \caption{\small Ablation study of loss components on the 360-AGD Hard Split. We investigate the impact of the  $\mathcal{L}_{\text{KL}}$, $\mathcal{L}_{\text{RTC}}$ and $\mathcal{L}_{\text{BCE}}$.}
  \label{tab:loss_ablation}
  \vskip-1ex
  \begin{tabular}{@{}ccc|ccc@{}}
    \toprule
    $\mathcal{L}_{\text{KL}}$ &
    $\mathcal{L}_{\text{RTC}}$ &
    $\mathcal{L}_{\text{BCE}}$ &
    \textbf{KLD $\downarrow$} &
    \textbf{SIM $\uparrow$} &
    \textbf{NSS $\uparrow$}  \\
    \midrule
 &            &   \checkmark    &1.596 &0.395  & 3.891 \\
   & \checkmark &    \checkmark     &1.459 & 0.442 &  4.374\\
\checkmark      &            & \checkmark &1.430  &0.450  &4.041  \\
\checkmark & \checkmark &            & 1.331  &\textbf{0.493}& 4.361 \\
\checkmark & \checkmark & \checkmark    &\textbf{1.306} &  0.474 &  \textbf{4.398}   \\
\bottomrule
  \end{tabular}
  \vskip-5ex
\end{table}

\textbf{Effects of Model Components.}
As shown in Table~\ref{tab:module_ablation}, each component provides incremental but distinct gains. The introduction of LoRA-based parameter-efficient adaptation yields an initial performance boost, demonstrating the necessity of task-specific feature tuning. Notably, the integration of Distortion-Aware Spectral Modulator (DASM) provides a critical correction for geometric warping, significantly reducing the KLD error. Furthermore, Omni-Spherical Densification Head (OSDH) complements this by refining spatial consistency, effectively restoring topological continuity from sparse activations. The full model achieves the best overall performance ($1.306$ KLD and $0.474$ SIM), confirming that our modular design effectively addresses the unique challenges of panoramic affordance prediction.

\textbf{Effects of Training Objectives.} Table~\ref{tab:loss_ablation} evaluates the contribution of each component within our multi-level training objective.
Training solely with pixel-level supervision $\mathcal{L}_{BCE}$ establishes a baseline but struggles to capture functional shapes from sparse annotations.
Incorporating $\mathcal{L}_{KL}$ introduces crucial distribution-level supervision, enhancing the model's capability to infer structured functional regions.
The addition of $\mathcal{L}_{RTC}$ further boosts performance, particularly on semantic-sensitive metrics such as SIM and NSS, by enforcing region-text alignment to critically disambiguate features under the one-shot setting. 
The full loss configuration yields the strongest overall gains, validating the synergistic effect of pixel-, distribution-, and semantic-level supervision.

\subsection{Hyperparameter Analysis}
We further analyze two key hyperparameters affecting model performance: the LoRA rank $r$ in the visual encoder and the number of seed points $k$ in the OSDH module.

\textbf{Impact of LoRA Adaptation Rank $r$.} As shown in Table~\ref{table:lora_rank} , model performance exhibits an approximately inverted-U trend as $r$ increases. 
When the rank is low ($r=4,8$), the representation capacity is insufficient to capture the complex nonlinear geometric distortions introduced by ERP projection, leading to suboptimal cross-domain alignment. 
Conversely, excessively high ranks ($r \geq 24$) result in noticeable performance degradation (\textit{e.g.}, KLD rises to $1.403$ at $r=32$). 
This suggests that over-parameterization may cause overfitting on the scarce one-shot samples or disrupt the robust semantic priors inherent in the pre-trained DINOv2 backbone. 
Our experiments indicate that $r=16$ achieves the optimal balance between adaptation flexibility and semantic preservation, which we adopt as our default configuration.

\textbf{Robustness to Top-$k$ Selection.} We further examine the sensitivity of the OSDH module to the top-$k$ value. 
Table~\ref{table:top_k} shows that our method exhibits remarkable stability across a wide range of top-$k \in [5,20]$: KLD fluctuates by only $0.006$ (from $1.306$ to $1.312$), while SIM remains nearly constant. 
This robustness stems from OSDH's integration of spherical self-similarity as a structural prior, coupled with a confidence-guided noise suppression mechanism, enabling reliable recovery of topologically continuous affordance regions from sparse activations without precise hyperparameter tuning.
Consequently, even when the top-$k$ varies, the model reliably recovers topologically continuous functional regions from sparse initial activations without fine-grained tuning of the seed threshold. 
This characteristic significantly enhances the practicality of our approach in real-world deployment.

\begin{table}[!t]
\centering
\begin{minipage}{0.48\textwidth}
    \centering
    \begin{minipage}[t]{0.48\linewidth}
        \centering
        \caption{\small Impact of varying LoRA Rank ($r$) on KLD, SIM, and NSS metrics.}
        \label{table:lora_rank}
        \begin{tabular}{lccc}
            \toprule
            $r$ & KLD $\downarrow$ & SIM $\uparrow$ & NSS $\uparrow$ \\
            \midrule
            4   &  1.343      &    0.465      &   4.296        \\
            8   &  1.341      &    0.467      &   4.285        \\
            \textbf{16} & \textbf{1.306} & \textbf{0.474} & \textbf{4.398} \\
            24  &  1.356      &    0.467      &   4.268        \\
            32  &  1.403      &    0.462      &   4.190        \\
            \bottomrule
        \end{tabular}
    \end{minipage}
    \hfill
    \setlength{\tabcolsep}{4.2pt}
    \begin{minipage}[t]{0.48\linewidth}
        \centering
        \caption{\small Impact of varying $top_k$ on KLD, SIM, and NSS metrics.}
        \label{table:top_k}
        \begin{tabular}{lccc}
            \toprule
            $top_k$ & KLD $\downarrow$ & SIM $\uparrow$ & NSS $\uparrow$ \\
            \midrule
            1   & 1.315  &0.468   & 4.352  \\
            5   & 1.308  & 0.474  & 4.381  \\
            \textbf{10} & \textbf{1.306} & \textbf{0.474} &4.398 \\
            20  &  1.312 & 0.473  & 4.413  \\
            30  &  1.320 & 0.470  & 4.413  \\
            \bottomrule
        \end{tabular}
    \end{minipage}
\end{minipage}
\vskip-3ex
\end{table}

\subsection{Performance in Real-World Scenarios}
\begin{figure}[!t]
\centering
\includegraphics[width=0.49\textwidth]{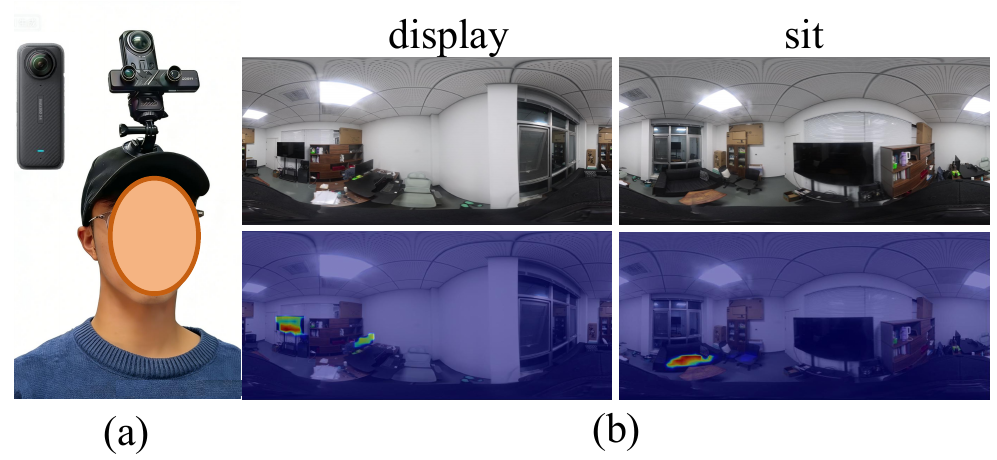}
\vskip-1ex
\caption{\small \textbf{Real-world evaluation.} (a) Wearable data collection setup. (b) Qualitative grounding results.}
\label{fig:real_world_val}
\vskip-4ex
\end{figure}

To further evaluate the generalization capability of PanoAffordanceNet in real-world unstructured environments, we conduct a series of field experiments. 
As illustrated in Fig.~\ref{fig:real_world_val}, we build a wearable data collection system by mounting an Insta360 X4 panoramic camera on a head-mounted cap to simulate the $360^{\circ}$ egocentric perspective of an embodied agent. In complex office and domestic scenarios, our model accurately localizes key functional regions (\textit{e.g.}, sit and display), despite varying illumination conditions and severe geometric distortions. The experimental results demonstrate that, by explicitly accounting for the geometric characteristics of $360^{\circ}$ imagery, our method achieves superior robustness in previously unseen real-world environments. This provides reliable functional priors for global decision-making and task planning in service robots.

\section{Conclusion}
This paper introduces a new task of holistic affordance grounding in $360^{\circ}$ indoor environments, shifting from isolated object-centric understanding to scene-level functional reasoning for embodied intelligence.
To address this task, we propose PanoAffordanceNet, an end-to-end framework that explicitly models panoramic distortion and sparse functional regions through distortion-aware spectral modulation and omni-spherical feature densification.
Jointly optimized with multi-level training objectives across pixel, distribution, and semantic layers, our method achieves robust and continuous functional grounding in complex $360^{\circ}$ environments.
Furthermore, we present 360-AGD, the first high-quality dataset for this task. 
We believe this work provides a solid foundation for embodied agents in real-world $360^{\circ}$ environments. 
Future research will explore temporal reasoning for dynamic scenes and cross-modal synergy with 3D spatial representations.

\bibliographystyle{IEEEtran}
\bibliography{IEEEabrv,main}

\end{document}